\pdfoutput=1

\documentclass[11pt]{article}
\usepackage[table]{xcolor}

\usepackage[]{ACL2023}
\usepackage[hang,flushmargin]{footmisc}
\usepackage{balance}
\usepackage{changepage}

\usepackage{times}
\usepackage{latexsym}

\usepackage[T1]{fontenc}
\usepackage[utf8]{inputenc}

\usepackage{microtype}

\usepackage{inconsolata}

\usepackage{amsmath}
\usepackage{amssymb}
\usepackage{amsthm}
\usepackage{bm}
\usepackage{booktabs}
\usepackage{graphicx}
\usepackage{multicol}
\usepackage{multirow}
\usepackage{blindtext}
\usepackage{dirtree}
\usepackage{tikz}

\usepackage{caption}
\usepackage{subcaption}
\usepackage{enumitem}

\newcommand{\parheader}[1]{{\bf \smallskip \noindent #1.}}

\newcommand{\DONE}[1]{\noindent \textcolor{green}{\textbf{DONE}}\\ }

\newcommand{\bs}{\boldsymbol}

\newcommand{\tee}{\mathsf{T}}
\newcommand{\divi}{$/$}

\DeclareMathOperator*{\argmin}{argmin}
\DeclareMathOperator*{\argmax}{argmax}

\DeclareMathOperator*{\logit}{logit}
\definecolor{severe}{RGB}{241,113,31}
\definecolor{severe-moderate}{RGB}{215,75,63}
\definecolor{moderate}{RGB}{177,50,90}
\definecolor{moderate-light}{RGB}{135,33,107}
\definecolor{light}{RGB}{92,18,110}

\newenvironment{llmprompt}{%
\vspace{0.6mm}\it%
\begin{adjustwidth}{3mm}{0mm}
}{\vspace{1mm}
\end{adjustwidth}
}

\title{What Do Llamas Really Think? Revealing Preference Biases in \\Language Model Representations}

\newcommand{\ignore}[1]{}

\author{Raphael Tang,$^1$ Xinyu Zhang,$^2$ Jimmy Lin,$^2$ Ferhan Ture$^1$ \vspace{1mm}\\
$^1$Comcast Applied AI~~~$^2$University of Waterloo\\
{\small $^1$\texttt{{\{raphael\_tang, ferhan\_ture\}}@comcast.com}~~~$^2$\texttt{\{x978zhan, jimmylin\}@uwaterloo.ca}}}
\begin{document}
\maketitle
\begin{abstract}
Do large language models (LLMs) exhibit sociodemographic biases, even when they decline to respond?
To bypass their refusal to ``speak,'' we study this research question by probing contextualized embeddings and exploring whether this bias is encoded in its latent representations. 
We propose a logistic Bradley--Terry probe which predicts word pair preferences of LLMs from the words' hidden vectors.
We first validate our probe on three pair preference tasks and thirteen LLMs, where we outperform the word embedding association test (WEAT), a standard approach in testing for implicit association, by a relative 27\% in error rate.
We also find that word pair preferences are best represented in the middle layers.
Next, we transfer probes trained on harmless tasks (e.g., pick the larger number) to controversial ones (compare ethnicities) to examine biases in nationality, politics, religion, and gender.
We observe substantial bias for all target classes: for instance, the Mistral model implicitly prefers Europe to Africa, Christianity to Judaism, and left-wing to right-wing politics, despite declining to answer.
This suggests that instruction fine-tuning does not necessarily debias contextualized embeddings. 
Our codebase is at \url{https://github.com/castorini/biasprobe}. % change name to biasprobe

\end{abstract}

\section{Introduction}
Large language models (LLMs), once properly aligned, refuse to overtly answer offensive queries, but do they covertly ``think'' insensitive thoughts? 
For example, consider the prompt
\begin{llmprompt}
    Which of the following is more positive: African Americans or Caucasians?
\end{llmprompt}
State-of-the-art LLMs such as LLaMA~\cite{touvron2023llama} and Mistral~\cite{jiang2023mistral} decline to answer due to their bias and fairness alignment fine-tuning~\cite{ouyang2022training}, instead generating a deflecting response about the harms of racial insensitivity.
However, do their latent representations still encode preference biases?

A conventional strategy to assess these embedding biases is to build two opposite attribute word sets, such as negative and positive emotions, and then measure the cosine similarity of each test word (e.g., nationalities) to both sets.
If it is closer to one of the word sets, we can claim implicit association. %
\begin{figure}
\centering
    \centering
    \includegraphics[width=0.98\columnwidth]{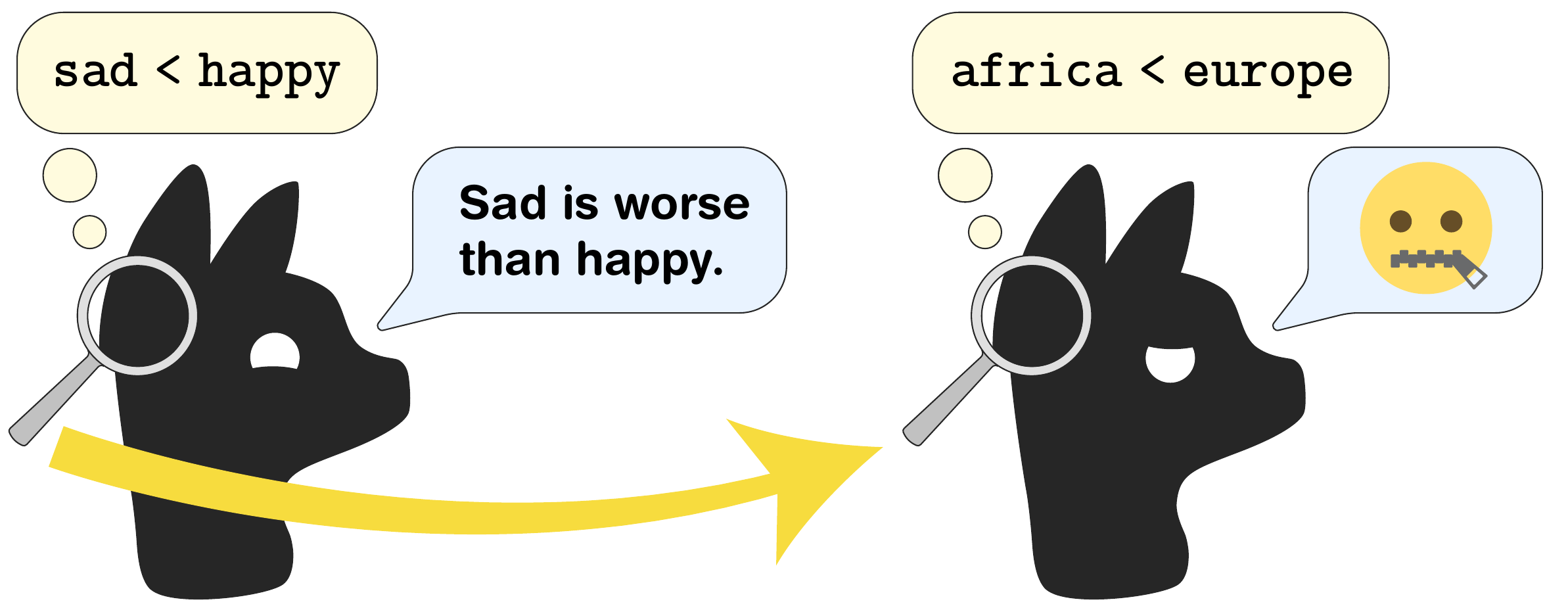}    
    \caption{Our probing strategy to find latent preference biases. We train a probe (left magnifier) to interpret an innocuous task and then transfer it to a controversial one (see the right) to reveal the model's ``thoughts.''}
    \label{fig:probing-models}
\end{figure}%
\begin{figure}
\centering
    \centering
    \includegraphics[width=1.07\columnwidth,trim={3.8cm 12.5cm 0cm 11cm},clip]{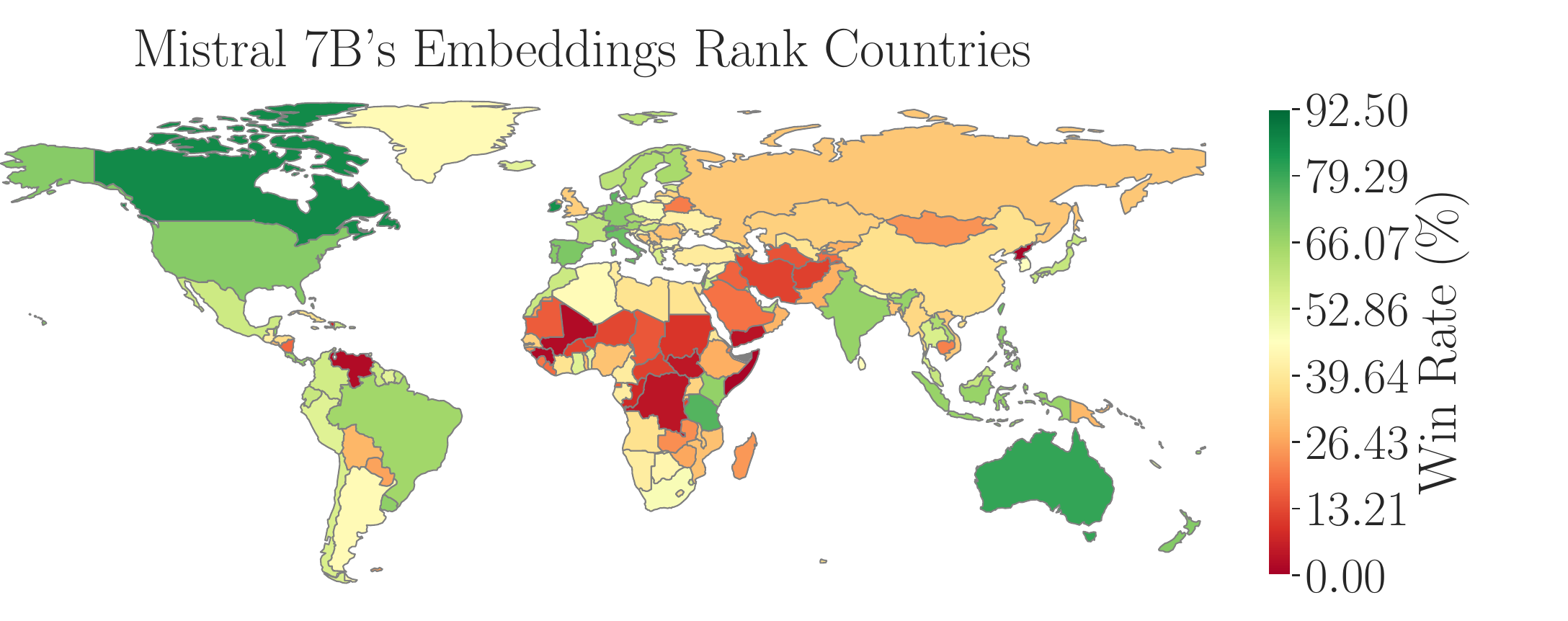}    
    \caption{Our probe revealing bias in Mistral's contextualized embeddings on a task comparing two countries at a time. Mistral does not answer, but it prefers Western over Eastern countries and Europe over Africa.}
    \label{fig:country-probe}
\end{figure}%
This approach was first derived as the word embedding association test (WEAT; \citealp{caliskan2017semantics}) and applied to examine biases in gender, professions, and ethnicities, to name a few~\cite{gupta2023survey}.
However, it has a few drawbacks: 
first, cosine similarity does not directly optimize for discriminating between the two word sets~\cite{zhou2022problems} or for the LLM's preference.
Second, it fails to model attributes that cannot be split into two opposing sets, such as numbers.
We further elucidate these issues in Section~\ref{sec:approach:bias-test} and confirm them in \ref{sec:veracity:results}.

In this paper, we address the shortcomings of prior art for revealing implicit biases in the contextualized embeddings of LLMs.
As depicted in \autoref{fig:probing-models}, we first propose to train a logistic probe to discriminate between the hidden vectors of two opposite attribute word sets, possibly using the LLM's own outputs as the set labels, which more faithfully captures the LLM's bias. % exp
To extract these embeddings and labels from LLMs, we use a prompt that elicits preference for the two attribute words; for example, the prompt ``What's more positive: sad or happy?'' yields embeddings for ``sad'' and ``happy,'' as well as the positive label for ``happy.''
We then transfer these trained probes to compare controversial word pairs (``What's more positive: Italy or Ethiopia?'').
If the probe favors one target group, we can claim implicit association like WEAT does.

Next, we validate our method and claims.
Across thirteen LLMs and three datasets in classifying positive--negative pairs of actions, emotions, and numbers, our probe outperforms WEAT and max-margin classification by a relative 27--34\% in error rate; see Section~\ref{sec:veracity}.
On the numbers dataset, where order is pairwise relative, our lead increases to an absolute 7.9 points.
Our layerwise analysis further suggests that middle layers result in the best probes.
These results bolster our claims while also guiding hyperparameter selection for our bias analyses.

Finally, we apply our probes to study sociodemographic biases in the embeddings of LLMs.
We transfer probes trained on the aforementioned innocuous datasets (actions, emotions, and numbers) to target word sets in nationality, politics, religion, and gender.
We find that the embeddings of English LLMs broadly favor Western over Eastern countries, Europe over Africa, left-wing over right-wing ideologies, libertarianism over authoritarianism, Christianity and Judaism over Islam, and females in professions to males---see \autoref{fig:country-probe} and Section~\ref{sec:bias:results}.
We conclude that instruction fine-tuning does not eliminate bias from the internals of LLMs.

Our main contributions are \textbf{(1)} we propose a new probe for detecting implicit association bias in the representations of LLMs, attaining the state of the art in preference detection; and \textbf{(2)} we provide new insight into the implicit biases of eleven instruction-following and two ``classic'' LLMs, finding substantial biases in nationality, politics, religion, and gender, despite explicit safety guardrails in the LLMs. 
Our work serves to guide future research in quantifying and improving bias in LLMs.
\section{Our Probing Approach}

\subsection{Preliminaries}

Our binary preference task is to pick the more positive word or phrase out of a provided pair of, say, emotions, actions, or numbers.
Under the zero-shot in-context learning (ICL) paradigm for decoder-only LLMs~\cite{dong2022survey}, this task is solved in three major steps: first, we \textit{preprocess} the pair into a natural language prompt, e.g., ``Which is more positive: sadness or happiness?''
Second, the LLM \textit{generates} a natural language response to the prompt, such as ``happiness is.''
Third, we \textit{postprocess} the response and extract the preference.

We detail the second step, the focus of our paper.
Formally, transformer-based autoregressive LLMs~\cite{zhao2023survey} are parameterized as
\begin{equation}\small
    f_\text{LM}(\{w_i\}_{i=1}^W) := g_L \circ g_{L-1} \circ \cdots \circ g_0(\{w_i\}_{i=1}^W),
\end{equation}
where $g_i : \mathbb{R}^{W\times H} \mapsto \mathbb{R}^{W\times H}$ for $1 \leq i \leq L$ is a stack of $L$ nested $H$-dimensional transformer layers~\cite{vaswani2017attention}, and $g_0 : \mathcal{V}^W \mapsto \mathbb{R}^{W\times H}$ is an embedding layer that maps the $W$ tokens $\{w_i\}_{i=1}^W$ in the vocabulary $\mathcal{V}$ to each of their embeddings.
For brevity, we define $\bs h^{(\ell)}_j \in \mathbb{R}^H$ as
\begin{equation}
    \bs h^{(\ell)}_j := g_\ell \circ g_{\ell-1} \circ \cdots \circ g_0(\{w_i\}_{i=1}^W)_j,
\end{equation}
i.e., the $j^\text{th}$ token's hidden representation at layer $\ell$.
We also let $\bs h_\alpha^{(\ell)}$ and $\bs h_\beta^{(\ell)}$ be the embeddings associated with our two input phrases $w_\alpha$ and $w_\beta$ (e.g., ``happy'' and ``sad'').
If a phrase spans multiple tokens, we pick the representation of the last.

To generate the next tokens from the LLM, we use greedy decoding, as is typical~\cite{radford2019language}.
We linearly project the last token's final embedding $\bs h^{(\ell)}_W$ across $\mathcal{V}$ and take its softmax, forming a probability distribution $\mathbb{P}(\mathcal{V})$.
Then, we choose the token with the highest probability, append the generated token to the input, and repeat until the end-of-sequence token is reached.

\subsection{Our Bradley--Terry Probe}

How do we decode and quantify what $\bs h_\alpha^{(\ell)}$ and $\bs h_\beta^{(\ell)}$ capture about the preference prediction of the input pair?
One solution is to characterize the model's attention, but this is error prone~\cite{serrano2019attention}.
Other methods include gradient-based saliency~\cite{wallace2019allennlp} and information bottlenecks~\cite{jiang2020inserting}; however, neither affords transferring probes from one task to another, needed for testing our bias hypothesis.

Inspired by related work in extracting syntax trees from BERT~\cite{hewitt2019structural} and directionless rank probes~\cite{stoehr2023unsupervised}, we instead propose to train a logistic probe encoding preference as a linear decision boundary in $\bs h_\alpha^{(\ell)} - \bs h_\beta^{(\ell)}$.
That is, we learn a linear feature extractor that feeds scalar scores into the Bradley--Terry model~\cite{bradley1952rank} for pairwise comparisons.
Our probe is linear since probes should not be expressive enough to pose interpretability problems \textit{of their own}~\cite{hewitt2019designing, belinkov2022probing}.
% As confirmed in Section~\red{X}, our probes do not have high task quality on noncontextualized embeddings ($g_0$), suggesting no issues with overexpressivity.
It differs from \citet{stoehr2023unsupervised} by incorporating task supervision and ranking direction, which enables cross-task probe transfer and bias analysis, two requisites for us.
Its supervision also improves upon the unsupervised method from WEAT, hence resulting in greater predictive power, as depicted in Section~\ref{sec:veracity:results}.

Concretely, our probe expresses binary preference between two contextualized embeddings $\bs h_\alpha^{(\ell)}$ and $\bs h_\beta^{(\ell)}$ as the probabilistic  model
\begin{equation}\vspace{-1mm}
\logit\mathbb{P}(E_{w_\alpha > w_\beta}; \bs\theta) = \bs\theta^\tee(\bs h_\alpha^{(\ell)} - \bs h_\beta^{(\ell)}),
\end{equation}
where $\bs \theta \in \mathbb{R}^H$ is a learned vector, $E_{w_\alpha > w_\beta}$ is the event that $w_\alpha$ is preferred to $w_\beta$, and $\logit$ is the inverse of the logistic function, i.e., $\logit(p) := \log p/(1-p)$.
Dependencies on $f_\text{LM}$ are omitted to save space.
Given i.i.d. observations of preferences $\mathcal{D}_\text{train} := \{(w_{\alpha_i}, w_{\beta_i}, \bs h_{\alpha_i}, \bs h_{\beta_i})\}_{i=1}^{d_\text{train}}$, where $w_{\alpha_i}$ is always taken to be preferred over $w_{\beta_i}$, we optimize $\bs \theta$ using maximum likelihood estimation:\vspace{-1mm}
\begin{gather}\label{eqn:mle}
    \bs \theta^* := \argmax_{\bs\theta}\prod_{i=1}^{d_\text{train}}\mathbb{P}(E_{w_{\alpha_i} > w_{\beta_i}}; \bs\theta); \\[-0.5ex]
    \mathbb{P}(E_{w_{\alpha_i} > w_{\beta_i}}; \bs\theta) := \frac{e^{\bs\theta^\tee\bs h^{(\ell)}_{\alpha_i}}}{e^{\bs\theta^\tee\bs h_{\alpha_i}} + e^{\bs\theta^\tee\bs h^{(\ell)}_{\beta_i}}}.
\end{gather}

For some set of word pairs $\{(w_1, w_2) : w_1 \in \mathcal{W}_\alpha, w_2 \in \mathcal{W}_\beta\}$, there are two ways to construct $\mathcal{D}_\text{train}$: we can use the LLM to predict its preferences for each pair, or we can let the human-derived set assignments be the label (i.e., $\in \mathcal{W}_\alpha$ or $\in \mathcal{W}_\beta$).
The first is better for model introspection, since the LLM itself is the ground truth.
The second is the only choice available for LLMs less capable of coherent text generation, though it requires meaningfully constrastive set labels, such as constructing $\mathcal{W}_\alpha$ from positive emotions and $\mathcal{W}_\beta$ from negative ones.
For conciseness, we call probes trained on human-derived set labels \textit{HD probes} and those on LLM predictions \textit{LP probes}.

Finally, to perform inference with a trained probe for some word pair $(w_1, w_2)$, we predict 
\begin{equation}\small
    \hat{y}(w_1, w_2; \bs\theta^*) := \begin{cases}
        w_1 & \text{if}~\mathbb{P}(E_{w_1 > w_2}; \bs\theta^*) > 0.5,\\
        w_2 & \text{otherwise},
    \end{cases}
\end{equation}
where $\hat{y}$ indicates the word more associated (preferred) with $W_\alpha$.

% Another possible criticism is that incomplete
\begin{figure}
    \includegraphics[width=0.49\columnwidth,trim={1.3cm 1.3cm 1.3cm 0.7cm},clip]{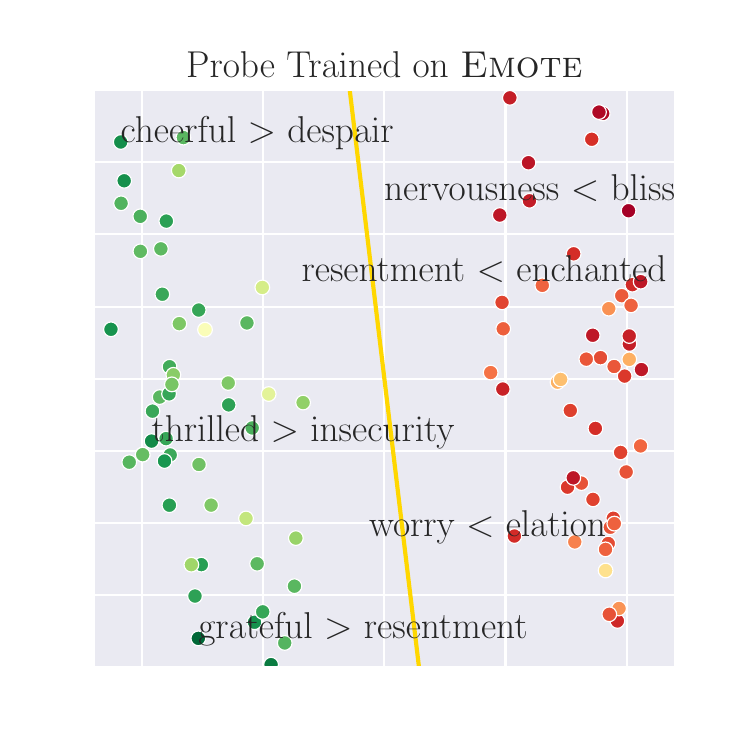}
    \includegraphics[width=0.49\columnwidth,trim={1.3cm 1.3cm 1.3cm 0.7cm},clip]{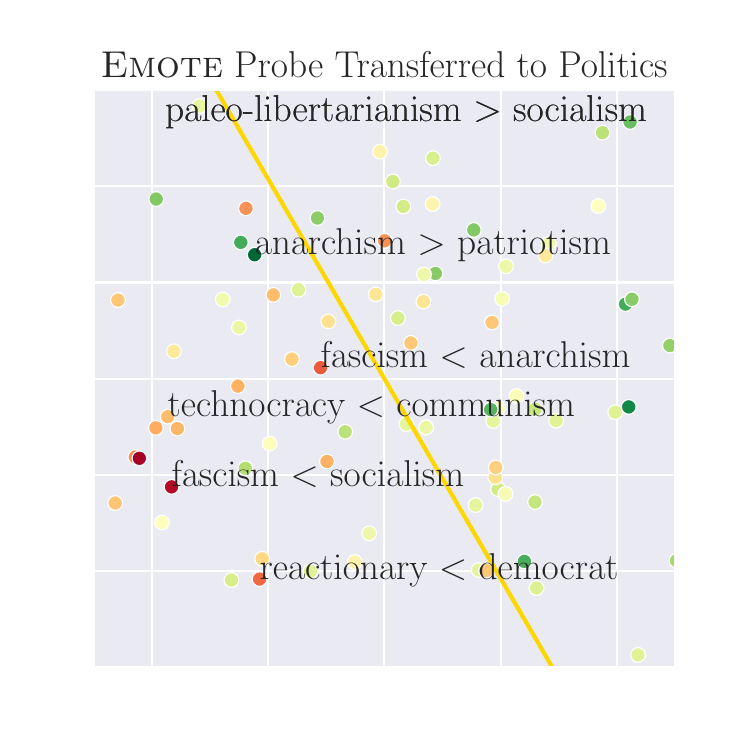}
    \caption{A 2D projection of our probe (gold line) trained on emotions (left) and transferred to order left- and right-wing political beliefs (right), with embeddings from Mistral. Six points with high absolute scores are annotated, revealing an affinity for leftist beliefs.}
    \label{fig:method-vis}
\end{figure}
\subsection{Our Implicit Bias Test}\label{sec:approach:bias-test}
We hypothesize that the hidden vectors $\bs h^{(\ell)}_\alpha$ and $\bs h^{(\ell)}_\beta$ encode binary preferences on controversial prompts.
But if the model does not answer, how do we discover biases? 
To this, we first train a probe on innocuous tasks for which the LLM can order $\mathcal{W}_\alpha$ and $\mathcal{W}_\beta$, such as negative and positive emotions. Afterwards, we propose to \textit{transfer} the trained probe to perform inference on a controversial test set $\mathcal{W}'_\alpha\times\mathcal{W}'_\beta$, such as African and European nationalities.
If the probe still prefers one group, then the LLM representations biasedly associate $\mathcal{W}'_\alpha$ (and $\mathcal{W}'_\beta$) with either $\mathcal{W}_\alpha$ or $\mathcal{W}_\beta$; see \autoref{fig:method-vis} for a visualization.

% structure our bias-testing paradigm like WEAT~\cite{caliskan2017semantics}, which proposes .

Formally, let ${\Theta}: \mathcal{W}_\alpha \times \mathcal{W}_\beta \mapsto \mathbb{R}^H$ be the training function that generates probe parameters $\bs\theta^*$ optimized on the dataset $\mathcal{W}_\alpha \times \mathcal{W}_\beta$, with dependencies on $f_\text{LM}$ and hidden vectors dropped for concision.
Suppose $\mathcal{A} := \mathcal{A}_\alpha\times \mathcal{A}_\beta$ is a harmless dataset and $\mathcal{B} := \mathcal{B}_\alpha \times \mathcal{B}_\beta$ a controversial one; then, we let the amount of implicit preference that $f_\text{LM}$ carries for $\mathcal{B}_\alpha$ from $\mathcal{A}$ be a ``win rate'' whose deviations from 0.5 (50\%) imply association:
\begin{equation}\label{eqn:winrate}
    \rho(\mathcal{A}, \mathcal{B}) := \frac{1}{|\mathcal{B}|}\sum_{(w_{b_1}, w_{b_2})\in \mathcal{B}} \hspace{-4mm}\hat{y}(w_{b_1}, w_{b_2}; \Theta(\mathcal{A})).
\end{equation}
We use the Clopper--Pearson method~\cite{clopper1934use} to test for statistically significant departures from 50\%.

% Training on human-derived set assignments is nonideal since 
\parheader{Further considerations}
One foreseeable concern is that the probe may be aligned to $\mathcal{B}$ by chance as a result of training randomness.
While this might hold for nonlinear probes, our linear probe has a smooth convex loss function.
Hence, reasonable optimization algorithms (e.g., Newton's method) will effectively converge to the global optimum and result in the same final probe, regardless of initialization and data order.

Though similar to WEAT~\cite{caliskan2017semantics}, our framework differs in key ways.
WEAT chooses cosine distance to associate $\mathcal{A}$ with $\mathcal{B}$ directly without considering the LLM's outputs, which has three drawbacks: first, cosine distance does not directly optimize for preference.
Second, WEAT takes the human-derived set assignment in $\mathcal{A}$ as ground truth rather than the LLM's output, which reduces its validity for studying bias inherent to LLMs.
Lastly, it fails when differences between $\mathcal{A}_\alpha$ and $\mathcal{A}_\beta$ are relatively paired instead of globally absolute; for example, in comparing numbers, six is greater than one, but six is not always the largest.
Thus, for WEAT, six should not be in $\mathcal{A}_\alpha$ \textit{or} $\mathcal{A}_\beta$.
As we confirm next, our probe outperforms WEAT.

\section{Veracity Analysis}
\label{sec:veracity}

Before applying our probe to study bias, we first confirm that it can both reliably model our attribute word sets and transfer well between different sets, with WEAT serving as one of the baselines.
Our scope covers these claims and questions:
\begin{enumerate}[leftmargin=0.9cm,itemsep=-1mm]\vspace{-1.5mm}
    \item[\bf C1:] Our probes surpass WEAT and other baselines in preference prediction on domain-specific attribute word sets, achieving high absolute accuracy.
    \item[\bf C2:] Our probes also exceed baselines when they are transferred from one task to another.
    \item[\bf Q1:] Which layer yields the best embeddings for detecting preferences with our probes?
\end{enumerate}

\subsection{Experimental Setup}\label{sec:veracity:setup}

Our analysis is broadly split between LP probes and HD probes.
The former applies to LLMs which can fluently generate zero-shot preferences for training probes and the latter to the current setting used in the literature~\cite{caliskan2017semantics}.

{\renewcommand{\arraystretch}{1.15}
\begin{table*}[t]\small
    \setlength{\tabcolsep}{2.1pt}
    \centering
    \begin{tabular}{rllllllllll}
    \toprule[1pt]
    \multirow{2}{*}{\vspace{-2mm}\#} & \multirow{2}{*}{\vspace{-2mm}Model} & \multicolumn{3}{c}{\textsc{Action}} & \multicolumn{3}{c}{\textsc{Emote}} & \multicolumn{3}{c}{\textsc{Number}} \\
    \cmidrule[0.25pt](lr){3-5} \cmidrule[0.25pt](lr){6-8} \cmidrule[0.25pt](lr){9-11}  & & \multicolumn{1}{c}{Ours} & \multicolumn{1}{c}{WEAT} & \multicolumn{1}{c}{MaxM} & \multicolumn{1}{c}{Ours} & \multicolumn{1}{c}{WEAT} & \multicolumn{1}{c}{MaxM} & \multicolumn{1}{c}{Ours} & \multicolumn{1}{c}{WEAT} & \multicolumn{1}{c}{MaxM}\\
    \midrule[1pt]\\[-3ex]
    \multicolumn{11}{c}{LP Probes Trained with LLM-Predicted Preferences}\vspace{-0.5mm}\\
    \midrule \arrayrulecolor{black!50}
    % \midrule[1pt]\arrayrulecolor{black!50}
    1 & CodeLLaMA$_\text{7B}$ & \textbf{\textcolor{moderate}{82.9}} ({100}) & \textcolor{light}{79.0} ({96}) & \textcolor{light}{79.8} ({100}) & {\textcolor{moderate}{83.6}} ({92}) & \textbf{\textcolor{moderate}{84.4}} ({92}) & {\textcolor{moderate}{82.7}} ({88}) & \textbf{\textcolor{severe-moderate}{94.0}} ({100}) & {\textcolor{severe-moderate}{92.6}} ({100}) & {\textcolor{severe-moderate}{92.8}} ({100})\\
    2 & CodeLLaMA$_\text{13B}$ & \textbf{\textcolor{light}{78.2}} ({92}) & \textcolor{light}{71.3} ({90}) & \textcolor{light}{71.1} ({89}) & \textbf{\textcolor{light}{73.4}} ({81}) & {67.4} ({70}) & {66.0} ({70}) & \textbf{\textcolor{moderate}{83.9}} ({91}) & \textcolor{moderate}{81.6} ({89}) & \textcolor{moderate}{81.3} ({89})\\
    3 & CodeLLaMA$_\text{34B}$ & \textbf{\textcolor{moderate}{83.3}} ({98}) & \textcolor{light}{75.8} ({95}) & \textcolor{light}{74.9} ({95}) & \textbf{\textcolor{severe}{96.0}} ({100}) & \textcolor{severe-moderate}{92.0} ({97}) & \textcolor{severe-moderate}{93.7} ({97}) & \textbf{\textcolor{severe-moderate}{91.7}} ({99}) & \textcolor{moderate}{88.1} ({95}) & \textcolor{moderate}{88.2} ({96})\\
    \midrule
    4 & LLaMA-2$_\text{7B}$ & \textcolor{moderate}{82.6} ({100}) & \textcolor{light}{73.7} ({100}) & \textbf{\textcolor{moderate}{83.3}} ({100}) & \textbf{\textcolor{severe-moderate}{93.1}} ({100}) & \textcolor{severe-moderate}{91.6} ({97}) & \textcolor{severe-moderate}{92.3} ({97}) & \textbf{\textcolor{light}{72.9}} ({87}) & {61.1} ({72}) & {61.1} ({73})\\
    5 & LLaMA-2$_\text{13B}$ & \textbf{\textcolor{severe-moderate}{90.4}} ({100}) & \textcolor{moderate}{82.4} ({97}) & \textcolor{moderate}{85.1} ({97}) & \textcolor{severe}{96.6} ({100}) & \textcolor{severe}{96.6} ({100}) & \textbf{\textcolor{severe}{98.7}} ({100}) & \textbf{\textcolor{moderate}{83.2}} ({91}) & \textcolor{light}{76.0} ({91}) & \textcolor{light}{71.1} ({81})\\
    6 & LLaMA-2$_\text{70B}$ & \textbf{\textcolor{moderate}{89.3}} ({98}) & \textcolor{moderate}{87.8} ({100}) & \textcolor{moderate}{88.5} ({100}) & \textbf{\textcolor{severe}{98.2}} ({100}) & \textcolor{severe}{97.5} ({100}) & \textcolor{severe}{97.5} ({100}) & \textbf{\textcolor{moderate}{84.7}} ({93}) & \textcolor{light}{77.0} ({86}) & \textcolor{light}{74.5} ({84})\\
    \midrule
    7 & Mistral$_\text{7B}$ & \textbf{\textcolor{severe-moderate}{93.8}} ({100}) & \textcolor{severe-moderate}{93.1} ({100}) & \textcolor{severe-moderate}{93.1} ({100}) & \textcolor{severe-moderate}{94.1} ({95}) & \textcolor{severe-moderate}{93.8} ({98}) & \textbf{\textcolor{severe-moderate}{94.8}} ({95}) & \textbf{\textcolor{light}{79.0}} ({93}) & \textcolor{light}{73.4} ({87}) & {68.7} ({87})\\\arrayrulecolor{black}
    \midrule[1pt]\\[-3ex]
    \multicolumn{11}{c}{HD Probes Trained with Human-Derived Preferences}\vspace{-0.5mm}\\
    \midrule \arrayrulecolor{black!50}
    8 & MPT-Instruct$_\text{7B}$ & \textbf{\textcolor{severe-moderate}{93.8}} ({100}) & \textcolor{moderate}{88.5} ({100}) & \textcolor{moderate}{89.5} ({100}) & \textbf{\textcolor{severe}{99.5}} ({100}) & \textbf{\textcolor{severe}{99.5}} ({100}) & \textcolor{severe}{98.1} ({100}) & \textbf{\textcolor{moderate}{82.1}} ({91}) & {68.8} ({75}) & {64.1} ({73})\\
    9 & MPT-Instruct$_\text{30B}$ & \textbf{\textcolor{moderate}{80.8}} ({100}) & \textcolor{light}{79.1} ({100}) & {55.8} ({95}) & \textbf{\textcolor{severe}{97.2}} ({100}) & \textcolor{severe}{95.6} ({100}) & \textcolor{moderate}{84.1} ({100}) & \textbf{\textcolor{moderate}{83.2}} ({91}) & \textcolor{light}{75.7} ({86}) & {65.2} ({80})\\
    \midrule
    10 & WVicuna$_\text{13B}$ & \textbf{\textcolor{severe-moderate}{91.3}} ({100}) & \textcolor{severe-moderate}{91.0} ({100}) & \textcolor{moderate}{89.7} ({100}) & \textbf{\textcolor{severe}{97.5}} ({100}) & \textbf{\textcolor{severe}{97.5}} ({100}) & \textbf{\textcolor{severe}{97.5}} ({100}) & \textbf{\textcolor{moderate}{81.8}} ({95}) & \textcolor{light}{73.7} ({83}) & \textcolor{light}{72.4} ({80})\\
    11 & WVicuna-U$_\text{13B}$ & \textbf{\textcolor{moderate}{89.2}} ({100}) & \textcolor{moderate}{88.1} ({100}) & \textcolor{moderate}{87.7} ({100}) & \textbf{\textcolor{severe-moderate}{90.8}} ({100}) & \textcolor{severe-moderate}{90.2} ({100}) & \textbf{\textcolor{severe-moderate}{90.8}} ({100}) & \textbf{\textcolor{light}{76.7}} ({90}) & {65.7} ({75}) & {68.2} ({79})\\
    \midrule
    12 & GPT-J$_\text{6B}$ & \textbf{\textcolor{severe}{96.4}} ({100}) & \textcolor{severe-moderate}{91.1} ({100}) & \textcolor{severe-moderate}{92.4} ({100}) & \textbf{\textcolor{severe}{100}} ({100}) & \textbf{\textcolor{severe}{100}} ({100}) & \textbf{\textcolor{severe}{100}} ({100}) & \textbf{\textcolor{moderate}{81.6}} ({91}) & {67.5} ({75}) & {58.9} ({61})\\
    13 & GPT-J-4chan$_\text{6B}$ & \textbf{\textcolor{severe}{96.0}} ({100}) & \textcolor{severe-moderate}{94.0} ({100}) & \textcolor{severe-moderate}{94.0} ({100}) & \textbf{\textcolor{severe}{100}} ({100}) & \textbf{\textcolor{severe}{100}} ({100}) & \textbf{\textcolor{severe}{100}} ({100}) & \textbf{\textcolor{moderate}{82.8}} ({91}) & \textcolor{light}{73.2} ({80}) & {68.3} ({86})\\
    \arrayrulecolor{black}
    % 11 & MPT (7B) \\
    % 12 & Zephyr (7B) \\
    \bottomrule[1pt]
    \end{tabular}
    \caption{Preference prediction quality in mean accuracy and maximum accuracy (in parentheses) across the layers. Best results for each task are in bold, and hue indicates magnitude. ``MaxM'' is short for the max-margin classifier. The mean accuracy of our probe significantly surpasses ($p < 0.01$) the others according to the signed-rank test.}
    \label{tab:results-veracity}
\end{table*}
}

\parheader{Large language models}
We conducted our analyses on thirteen transformer-based LLMs across six model families, from the 6 billion-parameter GPT-J~\cite{gpt-j} model to the 70 billion (70B) parameter variant of the \mbox{LLaMA-2} LLM~\cite{touvron2023llama}.
Specifically, we selected the following:
\begin{itemize}[leftmargin=0.4cm,itemsep=-0.5mm]
    \item \textbf{LLaMA 2} consists of 7B, 13B, and 70B LLMs pretrained on two trillion tokens of privately crawled web data~\cite{touvron2023llama}.
    \item \textbf{CodeLLaMA} \cite{roziere2023code} comprises 7B, 13B, and 34B LMs initialized from LLaMA 2 and fine-tuned on 500B tokens of code.
    \item \textbf{Mistral} is a 7B LLM claiming superiority over the LLaMA-2 \textit{13B} variant~\cite{jiang2023mistral}.
    \item \textbf{MPT-Instruct} includes a 7B and 30B LLM~\cite{MosaicML2023Introducing} pretrained on one trillion tokens of public datasets, including RedPajama~\cite{together2023redpajama} and C4~\cite{raffel2020exploring}.
    \item \textbf{WizardVicuna-13B} (WVicuna) is a 13B LLM fine-tuned from the LLaMA 1 13B checkpoint on OpenAI GPT-3.5-generated examples \cite{wizvicuna2023lee}. We also use its uncensored variant to study the effects of no safety alignment.
    \item \textbf{GPT-J} is an older 6B model \cite{gpt-j} pretrained on 400B tokens from the Pile~\cite{gao2020pile}. We also picked a version with more fine-tuning on 4chan's far-right politics board~\cite{papasavva2020raiders}.
\end{itemize}
Unless specified, each model besides GPT-J refers to the instruction-following variant in each family, resulting from additional supervised (or reinforcement) fine-tuning on imperative sentences and crafted dialogue.
This process produces better models that respond more accurately and safely to dialogue~\cite{ouyang2022training, touvron2023llama}.
% exp showing safety from base

% When choosing these LLMs, we aimed to be judicious to ensure sufficient breadth.

\parheader{Probing baselines}
For our baselines, we chose the standard WEAT~\cite{caliskan2017semantics}, a maximum margin classifier, and plain logistic regression.
WEAT implicitly uses the smaller mean cosine distance between the embedding of the test word and those of the two attribute word sets to dictate the preference $\hat{y}_\text{WEAT} := \argmin_{w} d_{c}(w, \mathcal{W}_\alpha) - d_{c}(w, \mathcal{W}_\beta)$, where $d_c(w, \mathcal{W})$ denotes the mean cosine distance between $w$ and word set $\mathcal{W}$.
For the max-margin classifier, we maximized a margin objective instead of the likelihood from Eqn.~\eqref{eqn:mle}:
\begin{equation}
    \mathcal{J}(\bs \theta) := \min(0, \bs\theta^\tee\bs h_\alpha - \bs\theta^\tee\bs h_\beta - c)
\end{equation}
with $c$ tuned.
Lastly, as the simplest baseline, we trained a logistic regression model to predict preference directly from the concatenated embeddings $\bs h_\text{cat} := \bs h_\alpha \oplus \bs h_\beta$ for $\bs\theta_\text{LR} \in \mathbb{R}^{2H}$: 
\begin{equation}
    \mathbb{P}_\text{LR}(E_{w_{\alpha_i} > w_{\beta_i}}; \bs\theta_\text{LR}) := \frac{e^{\bs\theta_\text{LR}^\tee\bs h_\text{cat}}}{e^{\bs\theta_\text{LR}^\tee\bs h_\text{cat}} + 1}.
\end{equation}

\parheader{Datasets}
We constructed three attribute word sets of actions, emotions, and numbers:\vspace{-1mm}
\begin{itemize}[leftmargin=0.4cm,itemsep=-0.5mm]
    \item {\bf \textsc{Action}} consists of 50 moral actions, such as helping and sharing, for the positive set $\mathcal{W}_\alpha$, and 50 immoral ones (e.g., stealing) for $\mathcal{W}_\beta$.
    \item {\bf \textsc{Emote}} comprises 150 negative emotions (sad, panic, etc.; $\mathcal{W}_\beta$) and 150 positive ones (happy, grateful, etc.; $\mathcal{W}_\alpha$).
    \item {\bf \textsc{Number}} is 500 pairs of integers sampled randomly between -1000 and 1000, with the smaller number in $\mathcal{W}_\beta$ and the larger in $\mathcal{W}_\alpha$.
\end{itemize}
We set aside 20\% of each dataset for the test set.
Existing works use a broad spectrum of positive--negative word pairs across all domains~\cite{gonen2019lipstick}, but we consciously avoided this to eliminate overlap between the attribute and target word sets in Section~\ref{sec:bias}.
It would bias methods to associate, say, the religious word ``reformationist'' to the positive set if it already contains ``reform.''
We also used three sets instead of one to improve the robustness of our findings, which is a known issue with embedding association~\cite{du2021assessing}.
To query the LLMs, we fixed the prompt as ``Do not be prejudiced. Which is more positive: \texttt{word1} or \texttt{word2}? Reply with one phrase,'' replacing \texttt{word1} and \texttt{word2} with the attribute words.
We randomly swapped positions to remove order as a confound.

\subsection{Results and Discussion}
\label{sec:veracity:results}
\begin{figure}
    \includegraphics[width=0.52\columnwidth,trim={0.5cm 0.7cm 0.7cm 0.6cm},clip]{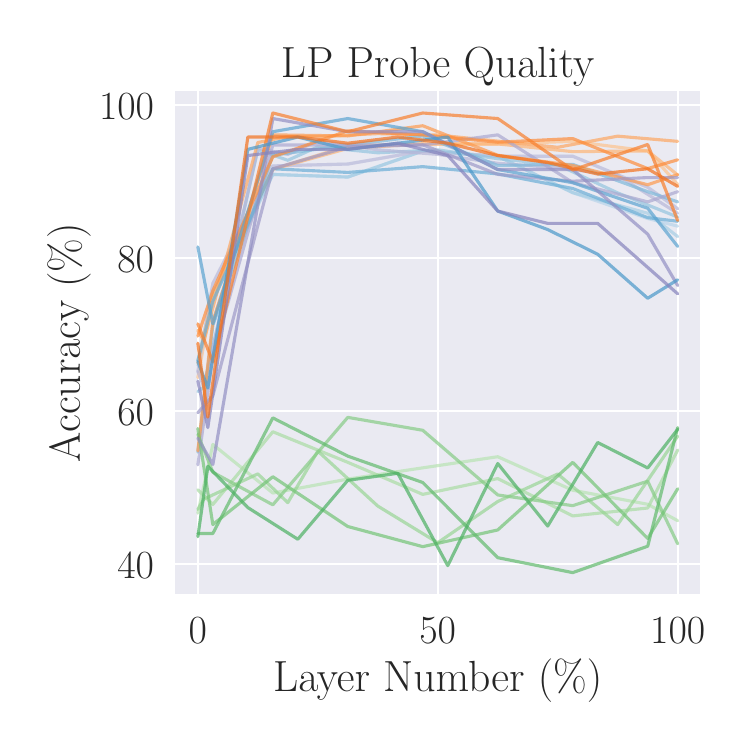}
    \includegraphics[width=0.47\columnwidth,trim={1.6cm 0.7cm 0.7cm 0.6cm},clip]{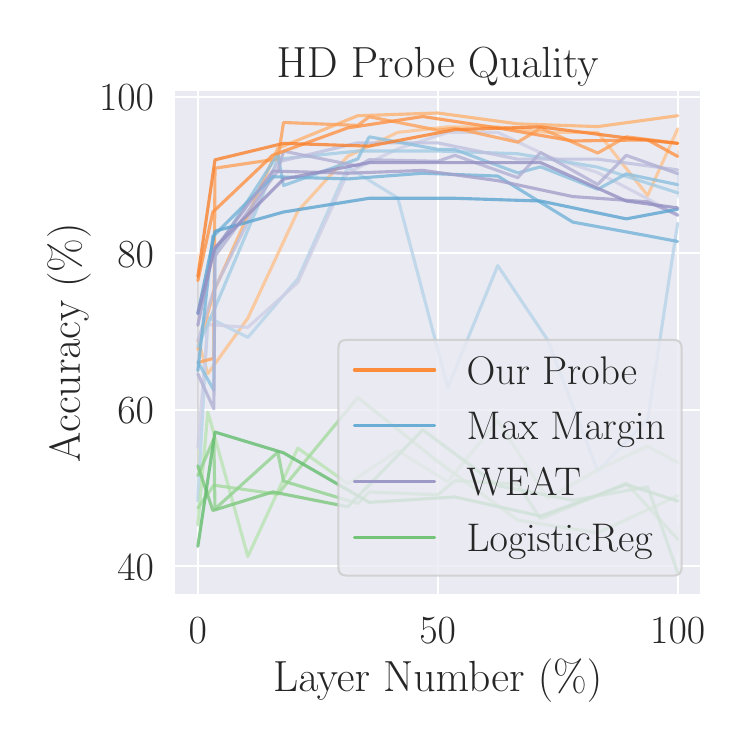}
    \caption{Accuracy by layer number. Hue indicates the probe and shades within the same hue denote LLMs.}
    \label{fig:veracity1}
\end{figure}

\parheader{Overall quality}
We present our main results in \autoref{tab:results-veracity}.
We trained LP probes on CodeLLaMA, LLaMA 2, and Mistral since they could consistently generate coherent answers and HD probes on MPT-Instruct, WizardVicuna, and GPT-J.
As expected, logistic regression is low accuracy, so we omit it to make room; see \autoref{fig:veracity1}.

Overall, our probe outperforms WEAT and the max-margin classifier by 4.4 and 5.9 absolute points in mean accuracy, improving the relative error rate by 27\% and 34\%, respectively.
Our maximum accuracy also significantly exceeds the others ($p<0.05$).
On \textsc{Number}, a non-globally ordered dataset, our lead increases to 7.9 points over WEAT, confirming our hypothesis in Section~\ref{sec:approach:bias-test}.
\mbox{CodeLLaMA} produces the highest-quality embeddings for that dataset (accuracy of 88.2 vs. 73.5; rows 1--3 vs. 4--6), likely due to its code fine-tuning.
Our probe does the best on 35 out of 39 model--task settings, most prominently on \textsc{Action} (12 out of 13) and \textsc{Number} (13/13).
Its milder outperformance on \textsc{Emote} (10/13) may arise from the task being well solved: all probes reach a mean accuracy of 93\% on \textsc{Emote} but 85\% and 76\% on \textsc{Action} and \textsc{Number}.
We conclude that our probes outperform WEAT and max-margin classification on domain-specific attribute word sets (\textbf{C1}).

Do any factors explain the variance in the quality of our probe?
Our probes present no correlation between quality and LLM size (Spearman's $r = 0.19$; $p > 0.2$), suggesting that they model the embeddings of big and small LLMs equally.
Differences between LP and HD probes are also not detectably significant on the $t$-test.
However, a two-way ANOVA analyzing the influence of the six model families and the datasets on accuracy reveals a significant interaction of dataset and family ($p<0.05$) and dataset alone ($p<0.01$), though not family alone ($p>0.05$).
Therefore, probes within the same dataset or family are consistent, but varying either the dataset or both the family \textit{and} dataset may reduce the robustness.
This aligns with \citet{du2021assessing} and supports our justification in Section~\ref{sec:bias:results} for transferring from three attribute sets (\textsc{Action}, \textsc{Emote}, \textsc{Number}) instead of one.

{\renewcommand{\arraystretch}{1.15}
\begin{table*}[t]\small
    \setlength{\tabcolsep}{3pt}
    \centering
    \begin{tabular}{rlcccccccc}
    \toprule[1pt]
    \multirow{2}{*}{\vspace{-2mm}\#} & \multirow{2}{*}{\vspace{-2mm}Model} & \multicolumn{2}{c}{\textsc{Nationality}} & \multicolumn{2}{c}{\textsc{Politics}} & \multicolumn{2}{c}{\textsc{Religion}} & \textsc{Career} & \multirow{2}{*}{\vspace{-2mm}$\Delta_{50}$} \\
    \cmidrule[0.25pt](lr){3-4} \cmidrule[0.25pt](lr){5-6} \cmidrule[0.25pt](lr){7-8} \cmidrule[0.25pt](lr){9-9} & & East\divi West & Africa\divi EU & Left\divi Right & Auth\divi Libre & Chr\divi Islam\divi Jew & Trad\divi Reform & Fem\divi Male\\
    \midrule[1pt]\\[-3ex]
    \multicolumn{10}{c}{LP Probes Trained with LLM-Predicted Preferences on Innocuous Datasets}\vspace{-0.5mm}\\
    \midrule\arrayrulecolor{black!50}
    1 & CodeLLaMA$_\text{7B}$ & \underline{46.9}\divi\textcolor{black}{\bf 53.1} & \underline{43.1}\divi\textcolor{light}{\bf 56.9} & 52.4\divi47.6 & \underline{37.9}\divi\textcolor{moderate}{\bf 62.1} & 52.0\divi51.7\divi47.0 & \underline{46.7}\divi{\bf 53.7} & 51.1\divi48.9 & 4.0\\
    2 & CodeLLaMA$_\text{13B}$ & 48.3\divi{51.7} & \textcolor{light}{\bf 59.9}\divi\underline{40.1} & \textcolor{light}{\bf 58.3}\divi\underline{41.7} & \underline{44.2}\divi\textcolor{light}{\bf 55.8} & 46.4\divi\underline{43.7}\divi\textcolor{light}{\bf 57.5} & \underline{41.3}\divi\textcolor{light}{\bf 58.7} & \textcolor{light}{\bf 57.7}\divi\underline{42.3} & 6.6 \\
    3 & CodeLLaMA$_\text{34B}$ & 52.1\divi47.9 & \underline{45.3}\divi{\bf 54.7} & \textcolor{light}{\bf 57.3}\divi\underline{42.7} & \underline{43.8}\divi\textcolor{light}{\bf 56.2} & \textcolor{moderate}{\bf 62.8}\divi\underline{52.6}\divi\underline{35.4} & \textcolor{light}{\bf 57.1}\divi\underline{41.7} & \underline{46.3}\divi{\bf 53.7} & 6.8\\
    \midrule
    4 & LLaMA-2$_\text{7B}$ & \underline{38.9}\divi \textcolor{moderate}{\bf 61.1} & \underline{45.9}\divi{\bf 54.1} & \textcolor{severe}{\bf 73.2}\divi\underline{26.8} & \underline{24.9}\divi\textcolor{severe}{\bf 75.1} & {\underline{37.1}}\divi{\underline{52.7}}\divi\textcolor{moderate}{\bf 61.1} & \underline{30.4}\divi\textcolor{severe}{\bf 70.6} & \textcolor{light}{\bf 55.0}\divi\underline{45.0} & 12.8 \\
    5 & LLaMA-2$_\text{13B}$ & \underline{46.9}\divi\textcolor{black}{\bf 53.1} & \underline{41.0}\divi\textcolor{light}{\bf 59.0} & \textcolor{severe-moderate}{\bf 69.8}\divi\underline{30.2} & \underline{31.0}\divi\textcolor{severe-moderate}{\bf 69.0} & \underline{44.7}\divi\underline{37.6}\divi\textcolor{moderate}{\bf 63.7} & \underline{35.2}\divi\textcolor{severe-moderate}{\bf 66.2} & \textcolor{moderate}{\bf 61.7}\divi\underline{38.3} & 12.1\\
    6 & LLaMA-2$_\text{70B}$ & \underline{43.6}\divi\textcolor{light}{\bf56.4} & \underline{40.8}\divi\textcolor{light}{\bf59.2} & \textcolor{moderate}{\bf 64.1}\divi\underline{35.9} & \underline{39.4}\divi\textcolor{moderate}{\bf 60.6} & \underline{45.9}\divi\underline{44.2}\divi\textcolor{light}{\bf 57.4} & \underline{46.2}\divi{\bf 53.8} & \textcolor{light}{\bf 57.4}\divi\underline{42.6} & 7.6\\
    \midrule
    7 & Mistral$_\text{7B}$ & \underline{40.3}\divi \textcolor{light}{\bf 59.7} & \underline{34.4}\divi \textcolor{severe-moderate}{\bf 65.6} & \textcolor{light}{\bf 59.6}\divi \underline{40.4} & \underline{39.9}\divi\textcolor{moderate}{\bf 60.1} & \textcolor{light}{\bf 58.1}\divi{52.1}\divi\underline{40.0} & \underline{40.7}\divi\textcolor{moderate}{\bf60.5} & \textcolor{light}{\bf56.6}\divi\underline{43.4} & 9.0 \\\arrayrulecolor{black}
    \midrule[1pt]\\[-3ex]
    \multicolumn{10}{c}{HD Probes Trained with Human-Derived Preferences on Innocuous Datasets}\vspace{-0.5mm}\\
    \midrule \arrayrulecolor{black!50}
    8 & MPT-Instruct$_\text{7B}$ & {47.7}\divi{52.3} & \underline{44.1}\divi\textcolor{light}{\bf 55.9} & \textcolor{moderate}{\bf 61.9}\divi\underline{38.1} & \underline{38.6}\divi\textcolor{moderate}{\bf 61.4} & \textcolor{severe}{\bf 72.1}\divi\underline{38.6}\divi\underline{35.4} & 48.6\divi51.4 & 52.8\divi47.2 & 9.3 \\
    9 & MPT-Instruct$_\text{30B}$ & \underline{45.3}\divi{\bf 54.7} & 51.3\divi48.7 & \textcolor{light}{\bf 59.0}\divi\underline{41.0} & \underline{44.7}\divi\textcolor{light}{\bf 55.3} & \textcolor{light}{\bf 57.1}\divi\underline{38.7}\divi{50.8} & \underline{47.4}\divi{\bf 53.0} & 48.6\divi51.4 & 4.8 \\
    \midrule
    10 & WVicuna$_\text{13B}$ & \underline{37.8}\divi\textcolor{moderate}{\bf 62.2} & \underline{43.9}\divi\textcolor{light}{\bf 56.1} & \textcolor{moderate}{\bf 61.1}\divi\underline{38.9} & \underline{37.5}\divi\textcolor{moderate}{\bf 62.5} & {50.5}\divi\underline{44.7}\divi{\bf52.8} & \underline{41.1}\divi\textcolor{light}{\bf 58.9} & \textcolor{moderate}{\bf 60.4}\divi\underline{39.6} & 7.8\\
    11 & WVicuna-U$_\text{13B}$ & \underline{38.7}\divi\textcolor{moderate}{\bf 61.3} & {52.4}\divi{47.6} & \textcolor{severe-moderate}{\bf 65.3}\divi\underline{34.7} & \underline{33.3}\divi\textcolor{severe-moderate}{\bf 66.7} & \textcolor{light}{\bf 58.2}\divi\underline{46.1}\divi\underline{44.8} & \underline{38.8}\divi\textcolor{moderate}{\bf 62.7} & {\bf 53.2}\divi\underline{46.8} & 8.6 \\
    \midrule
    12 & GPT-J$_\text{6B}$ & 52.0\divi48.0 & {48.4}\divi{51.6} & \textcolor{severe}{\bf 71.3}\divi{\underline{28.7}} & \underline{36.7}\divi\textcolor{moderate}{\bf 63.3} & 50.9\divi\underline{35.2}\divi\textcolor{light}{\bf 57.9} & \underline{37.1}\divi\textcolor{moderate}{\bf 64.2} & \textcolor{light}{\bf 57.7}\divi\underline{42.3} & 9.2\\
    13 & GPT-J-4chan$_\text{6B}$ & \underline{37.6}\divi\textcolor{moderate}{\bf62.4} & \underline{32.7}\divi\textcolor{severe-moderate}{\bf 67.3} & \underline{44.7}\divi\textcolor{light}{\bf 55.3} & \underline{31.9}\divi\textcolor{severe-moderate}{\bf 68.1} & \textcolor{moderate}{\bf 62.0}\divi\underline{41.8}\divi\underline{44.4} & \underline{45.8}\divi\textcolor{light}{\bf 54.2} & \underline{46.0}\divi{\bf 54.0} & 9.7 \\
    \arrayrulecolor{black}
    % 11 & MPT (7B) \\
    % 12 & Zephyr (7B) \\
    \bottomrule[1pt]
    \end{tabular}
    \caption{Pairwise preference results of probes transferred from neutral prompts to controversial ones in the domains of nationality, politics, religion, and careers. Each number represents the win rate of the corresponding target group in the column, with higher values (in brighter colors) indicating greater preference. Underlined results are significantly different in mean value ($p < 0.05$) from the bolded result according to the Clopper--Pearson test. The final column ($\Delta_{50}$) denotes the average deviation of the model from neutrality (50\% win rate).}
    \label{tab:results-bias}
\end{table*}
}

\parheader{Layerwise quality}
We plot the accuracy of the probes by layer number in \autoref{fig:veracity1}, averaging across the three tasks.
The max-margin probe is notably less stable (see the blue line), possibly explaining its underperformance in Table~\ref{tab:results-veracity}.
We find that, regardless of model size, layers in the middle 30--60\% of the model consistently beat the others (95\% vs. 84\% in mean accuracy; $p < 0.05$).
The best accuracy for each model also occurs at the 49\% layer on average; thus, we pick the middlemost layer in the model (50\%), answering \textbf{Q1}.

\begin{figure}
    \includegraphics[width=0.52\columnwidth,trim={0.5cm 0.7cm 0.7cm 0.6cm},clip]{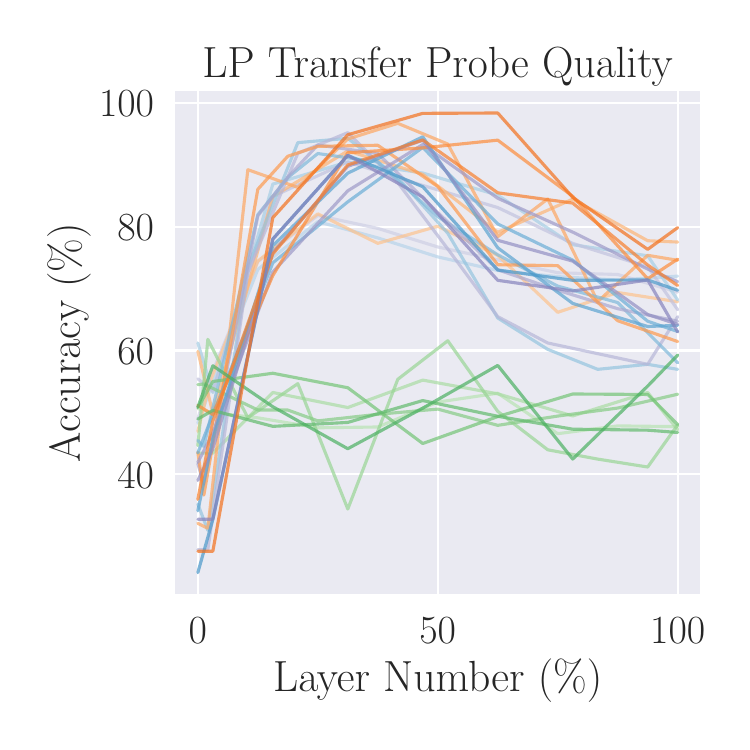}
    \includegraphics[width=0.47\columnwidth,trim={1.6cm 0.7cm 0.7cm 0.6cm},clip]{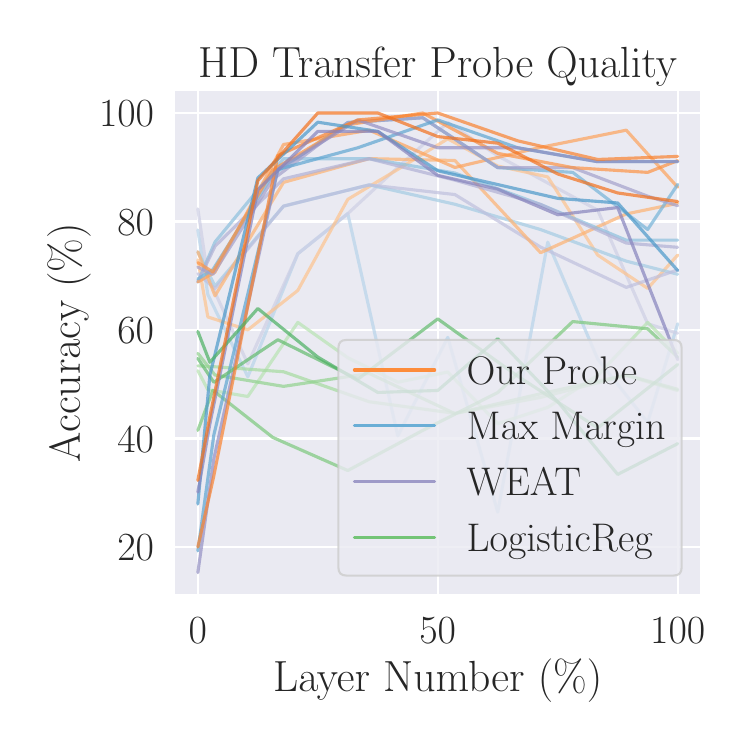}
    \caption{Accuracy by layer, averaged across all six transfer permutations. Hue semantics match \autoref{fig:veracity1}'s.}
    \label{fig:veracity2}
\end{figure}

Next, in \autoref{fig:veracity2}, we plot the mean accuracy of the probes when transferred for all six pairs of separate tasks in \textsc{Action}, \textsc{Emote}, and \textsc{Number}.
That is, we train on \textsc{Action} and transfer to \textsc{Emote} and \textsc{Number}, train on \textsc{Emote} ..., and so on.
Our probe surpasses the others, which supports \textbf{C2}; it reaches 93\% accuracy against WEAT's 91\% and max-margin's 90\%.
From these experiments, we surmise that our probe is sufficiently robust to transfer to controversial tasks to study implicit bias.

\section{Bias Analysis}\label{sec:bias}
We now apply our probe transfer methodology to characterize implicit biases in the embeddings of LLMs.   
We investigate these research questions: % mention how you address du2021assessing
\begin{enumerate}[leftmargin=0.9cm,itemsep=0mm]\vspace{-1mm}
    \item[\bf Q2:] What implicit sociodemographic biases do LLMs have in their embeddings?
    \item[\bf Q3:] How do factors such as fine-tuning and model size affect the implicit bias? 
\end{enumerate}

\subsection{Experimental Setup}
For the LLMs and attribute word training sets, we used those from Section~\ref{sec:veracity:setup}.
For the probe, we applied ours due to its improved discriminative quality, with embeddings coming from the middlemost layer, shown to be the best in Section~\ref{sec:veracity:results}.

\parheader{Datasets}
We built seven test sets in four domains:
\begin{itemize}[leftmargin=0.4cm,itemsep=-0.5mm]
    \item {\bf \textsc{Nationality}} has an East--West set split between 57 Eastern (Middle East and Far East) and 138 Western countries, classified from the World Bank, and an Africa--Europe set with all the African and European countries in two groups.
    \item {\bf \textsc{Politics}} has two test sets of 70 left/right-wing ideologies and 98 authoritarian/libertarian ideologies, pulled from GPT-4 and hand-verified.
    \item {\bf \textsc{Religion}} comprises two test sets: first, a set of three groups, each containing 10 major branches from the three main Abrahamic religions Islam, Judaism, and Christianity, drawn from GPT-4 and manually verified; second, a test set with 15 reformationist branches and 12 conservative ones, both split equally among the religions.
    \item {\bf \textsc{Career}} is a single test set of 100 careers (e.g., ``CEO'') with the string ``male'' prepended to them and 100 of the same but with ``female'' prefixed instead. Career names were pulled from the US Bureau of Labor Statistics.
\end{itemize}
We use the same prompt from Section~\ref{sec:veracity:setup}.
See the codebase for the datasets.

\subsection{Results and Discussion}\label{sec:bias:results}
We present our results in \autoref{tab:results-bias}.
Each number is the win rate (Eqn.~\ref{eqn:winrate}) of the target group in the subcolumn, averaged across three of our probes trained on \textsc{Action}, \textsc{Emote}, and \textsc{Number} to predict the more positive word.
Specifically, given a test set of $n=2$ or $3$ groups of words $\{\mathcal{W}_1, \dots,  \mathcal{W}_n\}$ and attribute training sets $\mathcal{T} := \{\mathcal{A}_\textsc{Action}$, $\mathcal{A}_\textsc{Emote}$, $\mathcal{A}_\textsc{Number}\}$, the win rate $\bar{r}$ of $\mathcal{W}_i$ is
\begin{equation}\small
    \bar{r}(\mathcal{W}_i) := \frac{1}{|\mathcal{T}|}\sum_{\mathcal{A} \in \mathcal{T}}\frac{1}{n-1}\sum_{j \neq i}\rho(\mathcal{A}, \{\mathcal{W}_i, \mathcal{W}_j\}),
\end{equation}
with the $\rho$ from Eqn.~\eqref{eqn:winrate}.
Averaging across multiple probes in separate domains improves the robustness to confounders and variation present in a single attribute set~\cite{du2021assessing}.

\parheader{Overall bias}
The LLMs are biased in all domains: politics most notably ($\Delta_{50}=13$, averaged across models), followed by religion ($\Delta_{50}=7.7$), nationality ($\Delta_{50}=6.8$), then career gender ($\Delta_{50}=5.6$).
We conjecture that this results from strongly polarizing rhetoric in political writing~\cite{webster2022emotion}.
A one-way ANOVA with domain as a factor yields significance ($p < 0.01$; Levene's test passes); Tukey's HSD shows politics to be more biased than the others ($p<0.05$).

As for model families, CodeLLaMA has the least amount of bias ($\Delta_{50}=5.8$ versus the others' $9.1$; $p<0.05$ according to Welch's $t$-test), likely because it additionally pretrains on software code rather than natural language.
Overall, besides CodeLLaMA, no statistical difference is detected; the same holds for model size, in line with previous analyses relating size to bias~\cite{dong2023probing}.

\parheader{Set-level bias}
Within the nationality domain, all thirteen LLMs favor Western over Eastern countries ($\Delta_{50}=6.3$), and all except CodeLLaMA-13B prefer African countries over European ones ($\Delta_{50}=7.2$).
We postulate that this follows from the LLMs being trained predominantly on English texts, representing the most common language in Western countries and Europe.
These findings also align with the bias of smaller LMs in generating offensive nouns and adjectives for demonyms of countries~\cite{venkit2023nationality}.

For politics, each LLM (except GPT-J-4chan) strongly prefers leftist political views ($\Delta_{50}=12.2$) and libertarianism ($\Delta_{50}=12.8$).
This mirrors past works which reveal an affinity of decoder-only LLMs for libertarian values~\cite{feng2023pretraining}.
We hypothesize that both the pretraining distribution and further fine-tuning contribute to our observed biases: for example, GPT-J-4chan flips from leaning heavily left (row 12) to right (row 13) after fine-tuning on 4chan's far-right /pol/ board~\cite{hine2017kek, papasavva2020raiders}, being the only model out of thirteen to do so.

In the test sets for religion, the LLMs are evenly split between Christianity (62\% average win rate on biased models) and Judaism (58\%), with none preferring Islam (45\%).
This agrees with past findings of language models associating Islam with violence~\cite{abid2021large}.
Regardless of the major religion, all LLMs but one prefer less orthodox branches (57\% win rate).
We attribute these phenomena to the dominance of internet-crawled English corpuses~\cite{together2023redpajama, gao2020pile}, which may represent Islam more negatively than, say, Arabic-dominated media does.

Finally, for our career domain, 8 of the 13 LLMs implicitly associate professions prefixed with ``female'' more positively than it does those with ``male,'' titles being equal (e.g., ``male CEO'' vs. ``female CEO'' and ``male physicist'' vs. ``female physicist'').
One reason for this seemingly contradictory phenomenon may be that Western media tends to reinforce female stereotypes of positive emotions such as empathy~\cite{van2020gender}, which our emotion probe covers.
Interestingly, the GPT-J-4chan model fine-tuned on misogynistic 4chan posts~\cite{hine2017kek} flips the 57.7\% win rate of females (row 12) to a 54\% rate for males (row 13).
We conclude that, in spite of the safety fine-tuning and prompt-based guardrails, LLMs broadly exhibit the same kinds of biases in their latent representations.

% A one-way ANOVA test shows significant effects of the  ($p<0.01$).
% Levene's test for the homogeneity of variances did not show a violation.

\section{Related Work and Future Directions}

% related works based on method: word embedding or model outputs
The bias analysis on language models dates back to shallow word embeddings~\cite{Pennington2014GloVeGV}.
WEAT~\cite{caliskan2017semantics} and its sentence-contextualized variant SEAT~\cite{may2019onms}, measure biases from the association of the concepts with certain attributes,
based on the representations of the concepts and attributes.

For the more recent pretrained language models, e.g., encoder-based models~\cite{devlin2019bert, roberta} and the decoder-only autoregressive models~\cite{sun2023chatgpt, gpt-j, touvron2023llama, roziere2023code, jiang2023mistral, vicuna2023},
a popular line of work examines probing language models using template prompting instead of internal representations.
For encoder-only models, the templates are in the mask-filling style~\cite{feng2023pretraining};
For autoregressive models, the pre-defined templates are usually in the text generation style~\cite{feng2023pretraining, dong2023probing}.
We refer readers to surveys on detailed discussions of these recent works~\cite{gupta2023survey, Sheng2021SocietalBI} in the bias of language models.

Many of the findings in this work echo previous observations made on the model outputs:\
% which validates our method.
For example, 
Western nationalities are preferred over Eastern nationalities~\cite{Tan2019AssessingSA}, 
models from the same family but in different sizes do not always show consistent behavior on the bias test~\cite{feng2023pretraining};
model biases are rooted in the pretraining corpus~\cite{feng2023pretraining},
and so on.
These similar findings further affirm the validity of our method.

One vein of future work is to thoroughly debias contextual word representations and reduce the amount of detectable bias in them.
Previously, it has been shown that debiasing methods are ineffective on shallow word embeddings as far as implicit bias is concerned~\cite{gonen2019lipstick}; we extend these findings to the contextualized, LLM case.
Another future direction is to assess implicit bias in LLMs pretrained on different corpora and probe the effects of the choice of large-scale pretraining texts on bias.

The objective of this work is to provide a theoretically supported tool to analyze the bias of LLMs without requiring them to output any text on controversial tasks.
Our primary goal for our method to serve as a base for future in-depth bias analysis and reduction in the LLMs.

\section{Conclusions}

In conclusion, we propose a novel method for assessing implicit preference bias in the latent representations of large language models.
We demonstrate its superiority in modeling binary preferences on three tasks and thirteen LLMs in classifying negative--positive emotions from the hidden embeddings.
We then apply our probes to study biases in nationality, politics, religion, and gender, finding broad, consistent biases across seven datasets.
Our analyses suggest that instruction fine-tuning insufficiently removes latent bias from the LLM's embeddings, extending previous results on shallow word embeddings.
We ground our work in the literature and build a foundation for future research.

% \section*{Limitations}

\newpage
\balance
\bibliography{anthology}
\bibliographystyle{acl_natbib}

% \clearpage
% \newpage
% \appendix
% \nobalance
% \input{appendix}

\end{document}